\documentclass{article}

\PassOptionsToPackage{numbers,compress}{natbib}

\usepackage[preprint]{neurips_2026}

\usepackage[utf8]{inputenc}
\usepackage[T1]{fontenc}
\usepackage{hyperref}
\usepackage{url}
\usepackage{booktabs}
\usepackage{amsmath}
\usepackage{amsfonts}
\usepackage{nicefrac}
\usepackage{microtype}
\usepackage{xcolor}
\usepackage{graphicx}
\usepackage{enumitem}
\usepackage{listings}
\usepackage{wrapfig}
\usepackage{tcolorbox}
\tcbuselibrary{listings, skins, breakable}
\newif\ifincludebibliography
\includebibliographytrue

\title{Teaching Language Models to Think in Code
}

\author{%
  Hyeon Hwang\\
  Korea University\\
  Seoul, South Korea \\
  \texttt{hyeon-hwang@korea.ac.kr}
  \And
  Jiwoo Lee\\
  Korea University\\
  Seoul, South Korea \\
  \texttt{hijiwoo7@korea.ac.kr}
  \And
  Jaewoo Kang\thanks{Corresponding author.}\\
  Korea University, AIGEN Sciences\\
  Seoul, South Korea \\
  \texttt{kangj@korea.ac.kr}
}

\begin{document}

\maketitle

\begin{abstract}
Tool-integrated reasoning (TIR) has emerged as a dominant paradigm for mathematical problem solving in language models, combining natural language (NL) reasoning with code execution. However, this interleaved setup has three key limitations: code often acts as a post-hoc verifier, intermediate NL computations are error-prone, and NL and code play overlapping rather than clearly distinct roles. We propose \textsc{ThinC} 
(Thinking in Code), a framework in which code itself serves as 
the reasoner rather than as a tool invoked by NL. A 
\textsc{ThinC} trajectory begins with a brief NL planning step, 
after which all reasoning unfolds through code blocks connected 
only by their execution outputs. We distill $12.2$k 
code-centric trajectories from a teacher model and train 
\textsc{ThinC}-1.7B and \textsc{ThinC}-4B with supervised 
fine-tuning followed by reinforcement learning. 
\textsc{ThinC}-4B consistently outperforms every TIR baseline 
on five competition-level math benchmarks and even surpasses 
the much larger Qwen3-235B-A22B-Thinking. Further analysis 
shows that \textsc{ThinC} reasons through code: $99.2\%$ of 
its final answers are grounded in interpreter output, and the 
model recovers reliably from code execution failures without 
intermediate NL reasoning. Our code and models will be released soon.

\end{abstract}

\section{Introduction}
\label{sec:introduction}

Recent advances in reinforcement learning (RL) over long chains of thought~\cite{wei2022chain} have substantially enhanced the mathematical reasoning capabilities of Large Language Models (LLMs), leading to powerful natural-language (NL) reasoners such as OpenAI o1~\cite{jaech2024openai} and DeepSeek-R1~\cite{guo2025deepseek}. Despite this progress, mathematical reasoning remains challenging for NL reasoners, particularly on problems requiring precise multi-step computation, where even a single arithmetic error can invalidate the entire reasoning process.

To make computation reliable, prior work has increasingly incorporated executable code into the reasoning process. Prompting-based approaches such as PAL~\cite{gao2023palprogramaidedlanguagemodels} and PoT~\cite{chen2023program} generate Python programs that solve mathematical problems end-to-end, delegating precise computation to a code interpreter. These methods demonstrated the reliability of code for mathematical computation and symbolic expression, but remain limited to single-pass program generation without iterative interaction with execution results. To combine the complementary strengths of NL reasoning and code execution, subsequent work introduced \textbf{tool-integrated reasoning (TIR)}~\cite{gou2024tora,wang2024mathcoder}, where NL handles high-level planning while code performs precise computation. TIR interleave NL reasoning with code execution over multiple turns, enabling iterative refinement and intermediate verification through interpreter feedback. Recent work has further expanded this paradigm along several directions. ReTool~\cite{feng2025retoolreinforcementlearningstrategic} uses RL to optimize tool-use strategies, ASTER~\cite{zhang2026asteragenticscalingtoolintegrated} emphasizes dense tool interaction throughout reasoning, and Tool-Star~\cite{dong2025toolstarempoweringllmbrainedmultitool} extends TIR to collaborative reasoning across multiple tools.

\begin{figure}[t]
    \centering
    \includegraphics[width=\linewidth]{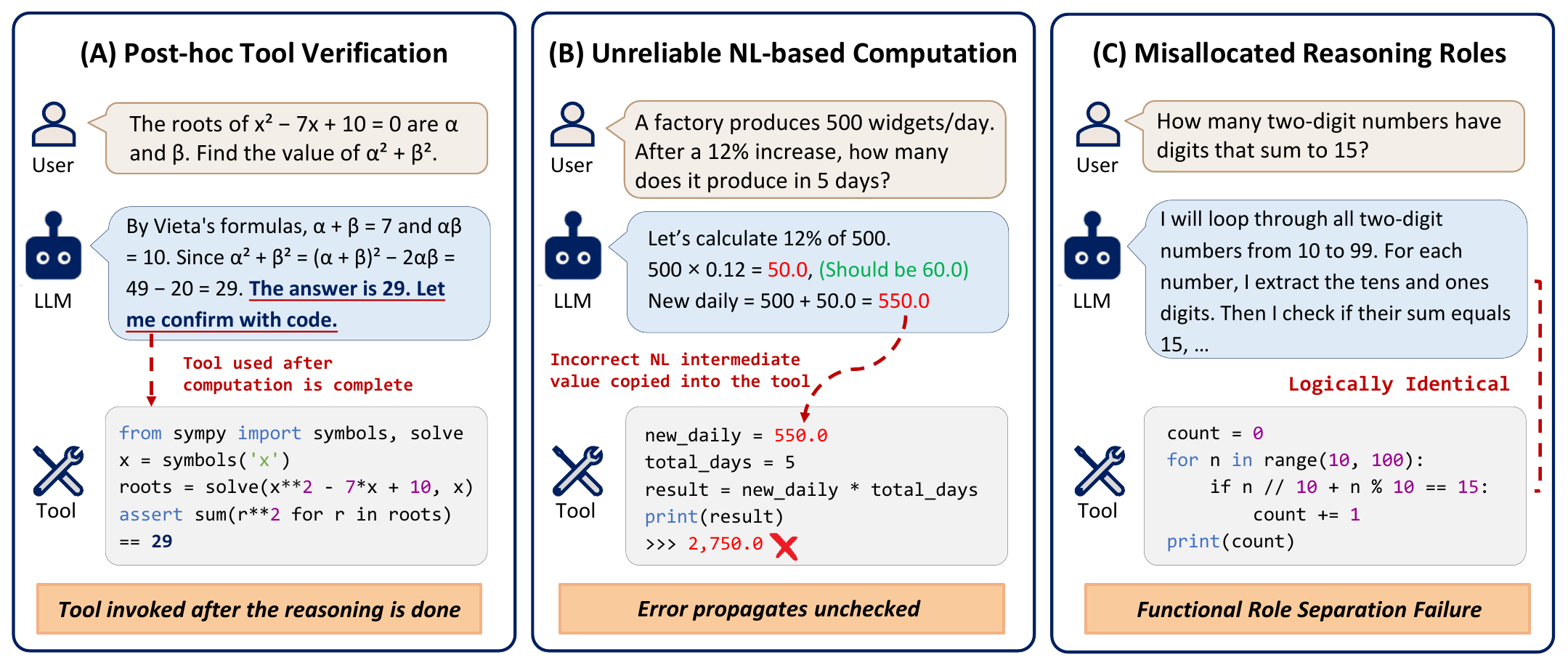}
    \caption{Three structural limitations of interleaved 
tool-integrated reasoning. \textbf{(A) Post-hoc tool 
verification}: the model completes a derivation in NL and runs 
code only to confirm the answer, so the interpreter performs no 
new computation. \textbf{(B) Unreliable NL-based computation}: an 
NL arithmetic error propagates silently into the next code block as a hard-coded 
constant. \textbf{(C) Misallocated Reasoning Roles}: the NL 
reasoning describes the very algorithm that the subsequent code 
re-implements.}
    \label{fig:motivation}
\end{figure}

However, as shown in Figure~\ref{fig:motivation}, TIR's interleaved reasoning paradigm suffers from 
three recurring structural limitations. \textbf{First}, the model often completes a derivation in NL 
first and then runs code only to confirm it; code becomes a 
post-hoc verifier rather than a reasoner, contributing no new 
computation. \textbf{Second}, when the model carries out arithmetic or algebraic 
steps in NL, a wrong value can be copied into the next code block as a 
hard-coded constant. The interpreter cannot detect the mistake, and the error silently affects the final answer.
\textbf{Third}, although NL reasoning excels at high-level 
planning and code can serve as a reasoner for precise 
mathematical expression and computation, interleaved TIR fails 
to separate these roles, leaving the two to do the same job. 
The NL reasoning lays out the algorithm step by step, taking 
on work that code is better suited for, while the code that 
follows merely transcribes the NL reasoning.

To address these limitations, we propose \textsc{ThinC} 
(Thinking in Code), a training framework in which code itself 
serves as the reasoner rather than as a tool driven by NL 
reasoning. A \textsc{ThinC} reasoning begins with a single 
brief planning step in NL that frames the strategy, after which 
all reasoning unfolds through code blocks connected only by 
their execution outputs. This structure resolves the three 
limitations by design: code performs derivations rather than 
verifying NL conclusions, every intermediate value is produced 
by the interpreter and therefore verified, and NL is restricted 
to high-level planning while code carries out all reasoning. 
We realize this paradigm in three stages: trajectory 
distillation from a teacher model via few-shot prompting to 
construct the $12.2$k \textsc{ThinC}-SFT dataset, supervised 
fine-tuning (SFT)  to establish the code-centric behavior prior, and 
RL with verifiable rewards~\cite{shao2024deepseekmath} to 
strengthen problem-solving.

We evaluate \textsc{ThinC} at two scales, \textsc{ThinC}-1.7B and 
\textsc{ThinC}-4B, built on Qwen3-1.7B and 
Qwen3-4B-Thinking-2507~\cite{yang2025qwen3technicalreport} 
respectively, across five competition-level math benchmarks (AIME 
2024--2026, HMMT 2025, and BeyondAIME~\cite{seed2025seed1}). 
\textsc{ThinC}-1.7B reaches $42.8\%$ average accuracy, exceeding 
Qwen3-1.7B by $10.6$ percentage points. \textsc{ThinC}-4B reaches 
$\mathbf{78.1\%}$, surpassing every TIR baseline in our 
evaluation and exceeding Qwen3-235B-A22B-Thinking, a much larger 
NL reasoner, on four of the five benchmarks. Further analysis 
shows that \textsc{ThinC}-4B reasons in a genuinely code-centric 
manner, with $\mathbf{99.2\%}$ of its final answers grounded in 
interpreter output rather than generated through NL reasoning. This 
behavior also makes \textsc{ThinC} robust when initial code 
executions fail, while interleaved TIR baselines degrade sharply.

\noindent\textbf{Our contributions are as follows.}
\begin{itemize}[leftmargin=1.5em,itemsep=2pt,topsep=2pt]
    \item We propose \textsc{ThinC}, a training framework that 
    teaches language models to treat code as the primary reasoner for mathematical problem solving rather than as a tool called by NL reasoning. \textsc{ThinC} consists of trajectory 
    distillation, SFT, and RL with verifiable rewards.

    \item We present the \textsc{ThinC}-SFT dataset of $12.2$k 
    code-centric trajectories, together with two trained models, 
    \textsc{ThinC}-1.7B and \textsc{ThinC}-4B. \textsc{ThinC}-4B reaches $\mathbf{78.1\%}$ average accuracy 
    across five competition-level math benchmarks, outperforming both all TIR baselines and the much larger Qwen3-235B-A22B-Thinking.

    \item We provide comprehensive analyses showing that \textsc{ThinC}-4B reasons in a genuinely code-centric 
    manner at inference time, and identify robustness to early 
    code execution failures as a concrete consequence of this 
    structure.
\end{itemize}

\section{Preliminaries}
\label{sec:preliminaries}
\subsection{Tool-Integrated Reasoning}
TIR augments a language model with one or more external tools that 
can be invoked during generation, such as code interpreters, search 
engines, or symbolic solvers. Solving a problem in TIR is a 
\emph{multi-turn} process: the model alternates between generating 
text and invoking tools, conditioning each subsequent action on the 
tool's output. In this work, we focus on the mathematical reasoning 
setting, where the tool is a Python interpreter $\mathcal{E}$ and 
each turn consists of a natural-language \emph{thought block} 
$t \in \mathcal{T}$, a \emph{code block} $c \in \mathcal{C}$ 
generated by the model, and an \emph{execution output} 
$o = \mathcal{E}(c)$ produced deterministically by the interpreter 
and appended to the context as a non-trainable observation. Given 
a problem $q$, the standard interleaved TIR paradigm produces 
trajectories of the form
\begin{equation}
\tau_{\mathrm{TIR}} = (q,\, t_1, c_1, o_1,\, t_2, c_2, o_2,\, \ldots,\, t_N, c_N, o_N,\, a),
\label{eq:tir}
\end{equation}
where $N$ is the number of turns and $a$ is the final answer.
All recent TIR systems~\cite{gou2024tora, feng2025retoolreinforcementlearningstrategic, zhang2026asteragenticscalingtoolintegrated} follow this structure.

\subsection{Supervised Fine-Tuning}
Supervised fine-tuning (SFT) adapts a pre-trained LLM to a target 
behavior by training on demonstration trajectories with a 
next-token prediction objective. In TIR, demonstrations are 
typically distilled from a stronger teacher model, and the choice 
of trajectories directly shapes the tool-use patterns that the 
model learns to produce~\citep{yu2025demystifyingreinforcementlearningagentic, 
feng2025retoolreinforcementlearningstrategic, 
zhang2026asteragenticscalingtoolintegrated}.
Given a dataset $\mathcal{D}_{\mathrm{SFT}}$ of trajectories, 
the SFT objective is
\begin{equation}
\mathcal{L}_{\mathrm{SFT}}(\theta) = 
- \mathbb{E}_{\tau \sim \mathcal{D}_{\mathrm{SFT}}}\!\left[
\sum_{k=1}^{|\tau|} m_k \log \pi_\theta(x_k \mid x_{<k})
\right],
\label{eq:sft}
\end{equation}
where $x_k$ is the $k$-th token of $\tau$ and 
$m_k \in \{0,1\}$ is a per-token loss mask. Prior TIR work 
commonly sets $m_k = 0$ 
for tool execution output tokens, restricting supervision to 
model-generated tokens only. We find no significant performance 
difference between the two choices and therefore use $m_k = 1$ 
for all tokens in this work.

\subsection{Reinforcement Learning with Verifiable Rewards}
For RL training in TIR, verifiable rewards are commonly 
used: each problem has a known ground-truth answer 
$a^\star(q)$, and a trajectory receives reward
\begin{equation}
r(\tau) = \mathbf{1}[a(\tau) = a^\star(q)],
\label{eq:reward}
\end{equation}
where $a(\tau)$ is the answer extracted from $\tau$. Exact-match 
verification removes the need for a learned reward model.

Group Relative Policy Optimization 
(GRPO)~\citep{shao2024deepseekmath} is a critic-free policy 
gradient algorithm widely used in this setting. For each problem 
$q$, GRPO samples a group of $G$ trajectories 
$\{\tau^{(g)}\}_{g=1}^{G}$ from the current policy $\pi_\theta$, 
and computes a group-relative advantage from their rewards:
\begin{equation}
A^{(g)} = \frac{r(\tau^{(g)}) - \mu_q}{\sigma_q}, \quad 
\mu_q = \frac{1}{G}\sum_{g=1}^{G} r(\tau^{(g)}), \quad 
\sigma_q^2 = \frac{1}{G}\sum_{g=1}^{G} \!\left(r(\tau^{(g)}) - \mu_q\right)^2.
\label{eq:advantage}
\end{equation}
Following DAPO~\citep{yu2025dapoopensourcellmreinforcement}, we adopt two modifications to 
the standard GRPO objective: \emph{token-level normalization} 
across the entire group rather than per-trajectory averaging, 
and \emph{asymmetric clipping} with 
$\epsilon_{\mathrm{low}} < \epsilon_{\mathrm{high}}$ that allows 
larger positive policy updates. The resulting clipped surrogate 
objective is
\begin{equation}
\mathcal{J}(\theta) = 
\mathbb{E}_{q,\,\{\tau^{(g)}\}}\!\left[
\frac{1}{\sum_{g} |\tau^{(g)}|}
\sum_{g=1}^{G} \sum_{k=1}^{|\tau^{(g)}|}
\min\!\left(\rho_k^{(g)} A^{(g)}, \,
\mathrm{clip}\!\left(\rho_k^{(g)}, 1{-}\epsilon_{\mathrm{low}}, 1{+}\epsilon_{\mathrm{high}}\right) A^{(g)}\right)
\right],
\label{eq:grpo}
\end{equation}
where $\rho_k^{(g)} = \pi_\theta(\tau_k^{(g)} \mid q, \tau_{<k}^{(g)}) / \pi_{\theta_{\mathrm{old}}}(\tau_k^{(g)} \mid q, \tau_{<k}^{(g)})$ 
is the per-token importance ratio.

\section{\textsc{ThinC}: Teaching Models to Think in Code}
\label{sec:method}

We present \textsc{ThinC}, a training framework that teaches 
language models to treat code itself as the reasoner for 
mathematical problem solving rather than as a tool invoked by natural language. \textsc{ThinC} consists of three components: (1) a code-centric trajectory format in which code itself serves as the reasoner (Section~\ref{sec:method:reasoning}); (2) a distillation and supervised fine-tuning procedure that induces this format in a student model (Section~\ref{sec:method:sft}); and (3) a multi-stage reinforcement learning procedure that further refines the resulting policy (Section~\ref{sec:method:rl}).

\subsection{\textsc{ThinC} Reasoning}
\label{sec:method:reasoning}

In interleaved TIR (Eq.~\ref{eq:tir}), an NL reasoner carries 
out the derivation and calls code as a tool. \textsc{ThinC} 
treats code itself as the reasoner. Code is a natural fit for 
this role because programming languages, like mathematics, are 
symbolic systems. Variables, operations, and functions in a 
program correspond directly to mathematical objects, allowing 
each reasoning step to be expressed and executed precisely.

A \textsc{ThinC} trajectory takes the form
\begin{equation}
\tau_{\mathrm{ThinC}} = (q,\, t_1,\, c_1, o_1,\, c_2, o_2,\, \ldots,\, c_N, o_N,\, a),
\label{eq:thinc}
\end{equation}
where $t_1$ is constrained to express \emph{strategy}, a 
high-level plan for solving the problem, rather than any 
step-by-step derivation of the answer. Unlike prior multi-turn 
TIR frameworks, which interleave thought and code at each step, 
our code-centric formulation uses a single initial thought $t_1$ 
to specify the overall solution strategy, and all subsequent 
reasoning is carried out through code. Each code block $c_i$ 
builds on the execution outputs of the preceding blocks, 
$o_1, \ldots, o_{i-1}$, and the final answer $a$ is obtained 
from the final execution output $o_N$.

\begin{figure}[t]
    \centering
    \includegraphics[width=\linewidth]{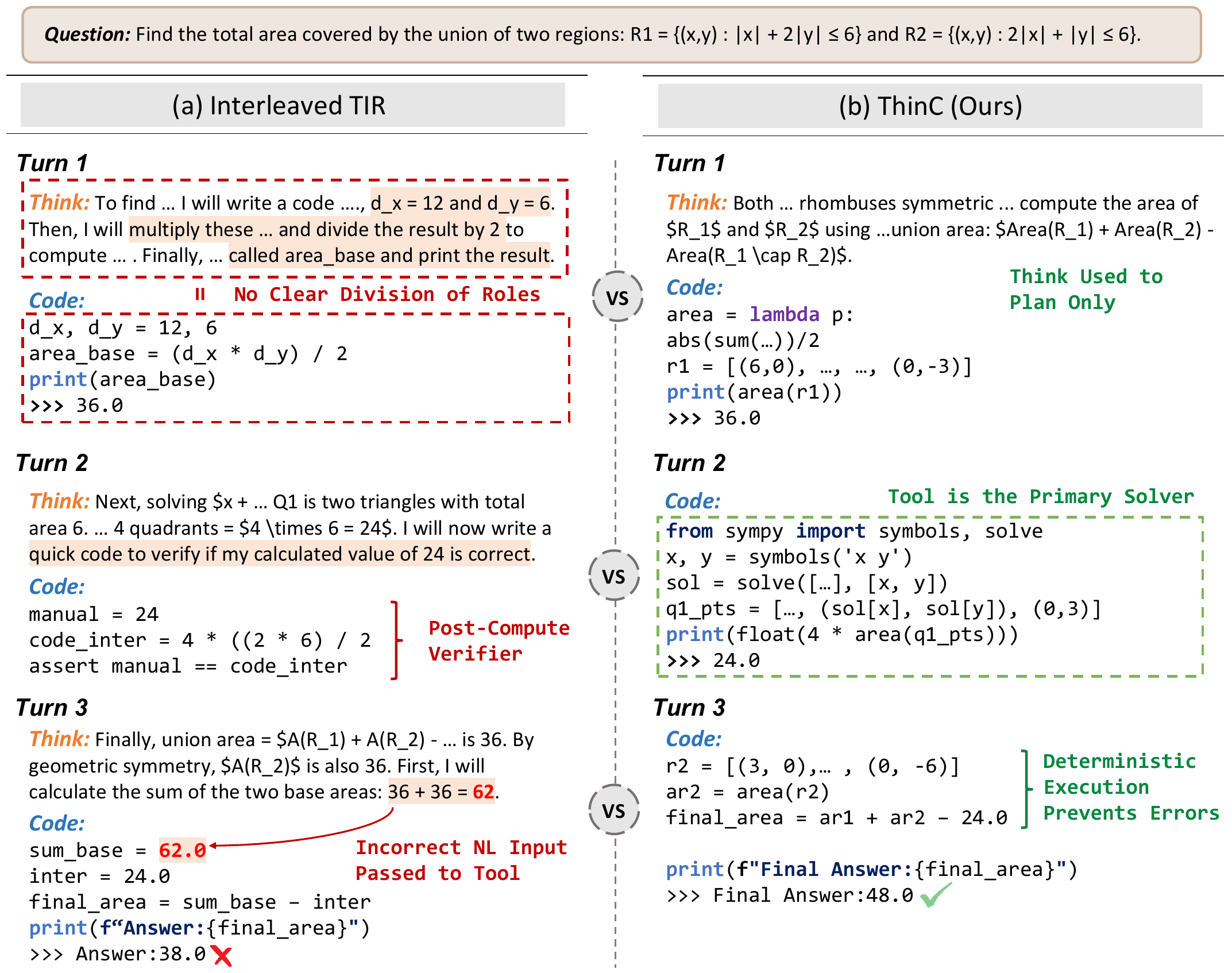}
    \caption{Comparison of interleaved TIR (left) and \textsc{ThinC} 
    (right). The three rows pair the structural 
    limitations from Section~\ref{sec:introduction} with how 
    \textsc{ThinC} avoids each. \textbf{(Top)} Interleaved TIR has NL describe the algorithm 
step by step before code re-implements that description, with 
NL and code doing the same job; \textsc{ThinC} restricts NL to 
high-level planning and lets code perform the derivation.
    \textbf{(Middle)} Interleaved TIR runs code as a post-hoc 
    verifier of an NL-derived answer; in \textsc{ThinC}, code is the 
    primary solver. \textbf{(Bottom)} An NL arithmetic error propagates into the next code 
    block as a hard-coded constant, leading interleaved TIR to the 
    wrong final answer; \textsc{ThinC} computes every 
    intermediate value through the interpreter and reaches the 
    correct answer.}
    \label{fig:thinc_overview}
    \label{fig:main}
\end{figure}

This simple structural change, illustrated in 
Figure~\ref{fig:thinc_overview}, resolves the three limitations 
of interleaved TIR identified in 
Section~\ref{sec:introduction} by construction:
\begin{itemize}[leftmargin=1.5em, itemsep=2pt, topsep=2pt]
    \item \textbf{Tool as a reasoner.} No thought block precedes 
    $c_i$ for $i \geq 2$, so each code block directly performs 
    a derivation step rather than acting as a post-hoc verifier, 
    making the interpreter an integral part of the reasoning 
    process.
    \item \textbf{Verified intermediates.} All intermediate 
    values are produced through the interpreter $\mathcal{E}$, 
    ensuring they are verified by construction and eliminating 
    unverified numerical computation in NL.
    \item \textbf{Specialized roles.} NL is restricted to 
    high-level planning in $t_1$, while code carries out all 
    subsequent reasoning, restoring the role separation that 
    interleaved TIR fails to maintain.
\end{itemize}
\subsection{Supervised Fine-tuning: Establishing Code-Centric Behavior}
\label{sec:method:sft}
To train models to reason through code, we distill \textsc{ThinC} 
trajectories from a strong teacher model and use them as 
supervised fine-tuning data.

Following prior 
work~\citep{zhang2026asteragenticscalingtoolintegrated}, we draw 
problems from Skywork-OR1~\citep{he2025skywork} and 
OpenMathReasoning~\citep{moshkov2025aimo}, restricted to 
English-language problems with positive integer answers. We 
sample one trajectory per problem from Qwen3.5-27B using a 
$3$-shot prompt that demonstrates the structure of 
Eq.~\ref{eq:thinc} (full prompt in Appendix~\ref{app:prompt}). 
We retain a distilled trajectory only if it (i) is correct, 
(ii) executes every code block without interpreter error, (iii) 
contains at least three code blocks, and (iv) spends less than 
$50\%$ of its tokens in the planning thought 
($|t_1|/|\tau| < 0.5$). Criteria (iii) and (iv) together enforce 
the code-centric structure of \textsc{ThinC} reasoning. 
Filtering yields the \textsc{ThinC}-SFT dataset of $12{,}200$ 
trajectories.

We fine-tune two base models, Qwen3-1.7B and 
Qwen3-4B-Thinking-2507, on \textsc{ThinC}-SFT using the SFT 
objective in Eq.~\ref{eq:sft}, with a context length of 32K, 
learning rate $7 \times 10^{-6}$ with cosine schedule, global 
batch size $16$, and $3$ epochs. We refer to the resulting 
checkpoints as \textsc{ThinC}-1.7B-SFT and 
\textsc{ThinC}-4B-SFT.

\subsection{Reinforcement Learning}
\label{sec:method:rl}

Starting from the SFT checkpoints, we further refine the policy 
using GRPO~\citep{shao2024deepseekmath} on 
DAPO-Math-17k~\citep{yu2025dapoopensourcellmreinforcement}. 
Following DAPO~\citep{yu2025dapoopensourcellmreinforcement}, we 
optimize the token-level policy gradient objective with 
Clip-Higher ($\epsilon_{\mathrm{low}} = 0.20$, 
$\epsilon_{\mathrm{high}} = 0.28$) and no KL divergence penalty, 
with a rollout prompt batch size of $128$ and $G = 8$ 
trajectories per prompt.

We train in three stages with increasing context budget; one 
epoch over the prompt set corresponds to roughly $140$ 
optimization steps. \emph{Stage 1} runs for $280$ steps (two 
epochs) on the full prompt set with a context length of 16K 
tokens and up to $20$ tool calls per trajectory. \emph{Stage 2} 
continues with the same context and tool budget but filters out 
problems whose Stage 1 policy already solves with $100\%$ pass 
rate, since these contribute zero group-relative advantage 
(Eq.~\ref{eq:advantage}); it runs for $120$ steps, ending at 
step $400$. \emph{Stage 3} begins at step $400$ with the same 
difficulty filtering, expanding the context to 32K and the tool 
budget to $40$ to allow longer trajectories on harder problems. 
We refer to the final checkpoints as \textsc{ThinC}-1.7B and 
\textsc{ThinC}-4B.

\section{Experiments}
\label{sec:experiments}
\subsection{Experimental Setup}
\label{sec:exp:setup}
\begin{table*}[t]
  \centering
  \caption{Comparison on AIME 2024, AIME 2025, AIME 2026, HMMT 
  2025 February, BeyondAIME, and their arithmetic mean. Scores 
  are reported as avg@$16$ (mean per-problem accuracy averaged 
  across problems) under a $32$K-token inference budget. 
  Smaller numbers indicate 95\% confidence intervals. Asterisk 
  ($^{*}$) denotes models prompted to use the Python 
  interpreter without further training. Qwen3.5-27B is the 
  teacher model used for distillation, evaluated with the 
  same $3$-shot prompt used during \textsc{ThinC}-SFT dataset 
  generation. \textbf{Bold}: best; \underline{underline}: 
  second-best.}
  \label{tab:main}
  \footnotesize  
  \setlength{\tabcolsep}{4pt}
  \renewcommand{\arraystretch}{1.1}
  \newcommand{\ci}[1]{\,{\scriptsize $\pm #1$}}  
  \begin{tabular*}{\textwidth}{@{\extracolsep{\fill}}lcccccc}  
    \toprule
    Model & AIME24 & AIME25 & AIME26 & HMMT25 & BeyondAIME & Average \\
    \midrule
    \multicolumn{7}{c}{\textit{NL-Only Reasoning Models}} \\
    \midrule
    Qwen3-1.7B                          & $47.3$\ci{13.3} & $35.0$\ci{14.2} & $36.2$\ci{14.0} & $22.5$\ci{12.6} & $19.8$\ci{6.0} & $32.2$\ci{5.5} \\
    Qwen3-4B-Thinking-2507              & $79.2$\ci{11.9} & $73.1$\ci{13.5} & $76.7$\ci{12.9} & $50.2$\ci{15.6} & $45.8$\ci{8.6} & $65.0$\ci{5.7} \\
    OpenReasoning-Nemotron-7B           & $76.9$\ci{11.8} & $73.1$\ci{13.0} & $72.1$\ci{13.3} & $52.7$\ci{14.4} & $38.6$\ci{7.7} & $62.7$\ci{5.5} \\
    gpt-oss-20B (medium)                & $79.4$\ci{11.5} & $69.8$\ci{12.0} & $76.9$\ci{10.7} & $51.0$\ci{13.9} & $46.2$\ci{7.6} & $64.7$\ci{5.1} \\
    Qwen3-235B-A22B-Thinking            & $\mathbf{90.6}$\ci{9.8} & $80.6$\ci{12.3} & $\underline{82.1}$\ci{11.5} & $68.8$\ci{14.2} & $\underline{54.1}$\ci{8.6} & $\underline{75.2}$\ci{5.1} \\
    \midrule
    \multicolumn{7}{c}{\textit{Tool-Integrated Reasoning Models}} \\
    \midrule
    Qwen3-1.7B*                         & $42.1$\ci{9.1}  & $36.2$\ci{15.1} & $32.1$\ci{13.3} & $19.6$\ci{12.6} & $18.9$\ci{6.0} & $29.8$\ci{5.2} \\
    Qwen3-4B-Thinking-2507*             & $77.3$\ci{8.8}  & $69.6$\ci{14.5} & $72.3$\ci{13.6} & $46.0$\ci{15.5} & $47.8$\ci{8.8} & $62.6$\ci{5.6} \\
    CoRT-1.5B                           & $35.8$\ci{12.0} & $30.0$\ci{13.8} & $25.8$\ci{11.2} & $23.1$\ci{11.6} & $13.8$\ci{4.6} & $25.7$\ci{5.0} \\
    DemyAgent-4B                        & $71.5$\ci{13.8} & $71.5$\ci{13.9} & $65.0$\ci{13.2} & $41.5$\ci{14.9} & $37.1$\ci{7.7} & $57.3$\ci{5.8} \\
    ASTER-4B                            & $78.8$\ci{11.5} & $\underline{84.6}$\ci{10.2} & $78.8$\ci{11.0} & $\underline{73.1}$\ci{11.8} & $54.0$\ci{8.1} & $73.8$\ci{4.7} \\
    rStar2-Agent-14B                    & $79.0$\ci{12.5} & $68.5$\ci{14.0} & $72.3$\ci{13.7} & $50.4$\ci{14.7} & $37.4$\ci{7.9} & $61.5$\ci{5.7} \\
    ReTool-32B                          & $65.8$\ci{14.1} & $56.0$\ci{14.6} & $52.7$\ci{14.6} & $47.1$\ci{14.1} & $31.8$\ci{7.6} & $50.7$\ci{5.9} \\
    \midrule
    \multicolumn{7}{c}{\textit{\textsc{ThinC} Reasoning Models}} \\
    \midrule
    \multicolumn{7}{l}{\textit{Teacher Model}} \\
    Qwen3.5-27B*                        & $74.8$\ci{10.5} & $64.0$\ci{11.0} & $75.6$\ci{10.2} & $63.3$\ci{12.9} & $45.7$\ci{7.5} & $64.7$\ci{4.7} \\
    \multicolumn{7}{l}{\textit{Student Models}} \\
    \textsc{ThinC}-1.7B                 & $59.0$\ci{9.2}  & $50.2$\ci{14.9} & $42.9$\ci{13.7} & $39.0$\ci{14.5} & $22.7$\ci{6.6} & $42.8$\ci{5.5} \\
    \textsc{ThinC}-4B                   & $\underline{88.3}$\ci{8.8} & $\mathbf{85.8}$\ci{8.8} & $\mathbf{86.0}$\ci{9.8} & $\mathbf{74.0}$\ci{12.7} & $\mathbf{56.1}$\ci{8.1} & $\mathbf{78.1}$\ci{4.4} \\
    \bottomrule
  \end{tabular*}
\end{table*}

\paragraph{Benchmarks.}
We evaluate on five competition-level mathematical reasoning 
benchmarks: AIME 2024, AIME 2025, AIME 2026, 
HMMT 2025 February, and BeyondAIME~\citep{seed2025seed1}.

\paragraph{Baselines.}
We compare \textsc{ThinC} to two groups of baselines. The first 
is \emph{NL-only reasoners}: Qwen3-1.7B and 
Qwen3-4B-Thinking-2507~\citep{yang2025qwen3technicalreport} (our base models), 
OpenReasoning-Nemotron-7B~\citep{ahmad2025opencodereasoningadvancingdatadistillation}, 
gpt-oss-20B~\citep{openai2025gptoss120bgptoss20bmodel}, and 
Qwen3-235B-A22B-Thinking~\citep{yang2025qwen3technicalreport}. The second is 
\emph{tool-integrated reasoners}: 
CoRT-1.5B~\citep{li2025teachinglanguagemodelsreasoncort}, 
DemyAgent-4B~\citep{yu2025demystifyingreinforcementlearningagentic}, 
ASTER-4B~\citep{zhang2026asteragenticscalingtoolintegrated}, 
rStar2-Agent-14B~\citep{shang2025rstar2agentagenticreasoningtechnical}, and 
ReTool-32B~\citep{feng2025retoolreinforcementlearningstrategic}. 
We additionally evaluate Qwen3-1.7B and Qwen3-4B-Thinking-2507, our base models prompted to use the Python interpreter without additional training, to separate the effect of \textsc{ThinC} training from those of the underlying base model and the tool-use prompt. We also report results for Qwen3.5-27B with our $3$-shot demonstration, as this model is used as the teacher for trajectory distillation.

\paragraph{Evaluation Protocol.}
For each benchmark, we sample $16$ trajectories per problem under 
a $32$K-token inference budget and report the average accuracy 
(avg@$16$). We sample \textsc{ThinC} with temperature $0.6$ and top-$p$ 
$1.0$, and follow the sampling parameters recommended by each 
baseline's original publication or official release. All models are evaluated 
with a Python interpreter providing access to the standard 
library (including \texttt{itertools} and \texttt{collections}) 
and the scientific computing libraries \texttt{numpy}, 
\texttt{scipy}, and \texttt{sympy}. All baselines are run in 
the same environment.
\subsection{Main Results}
\label{sec:exp:main}

\paragraph{\textsc{ThinC} delivers consistent gains at both scales.}
As shown in Table~\ref{tab:main}, \textsc{ThinC}-4B achieves the strongest overall result, with an average score of $78.1\%$ and the best performance on four of the five benchmarks. It outperforms all tool-integrated reasoning baselines, including substantially larger systems such as rStar2-Agent-14B and ReTool-32B. In addition, it surpasses Qwen3-235B-A22B-Thinking, the strongest open-source NL-only reasoner in our comparison, by $2.9$ points on average. The advantage is particularly large on the more challenging benchmarks, HMMT25 and BeyondAIME. Remarkably, \textsc{ThinC}-4B also exceeds its distillation teacher, Qwen3.5-27B with our $3$-shot demonstration, on all five benchmarks by $13.4$ points on average, despite being much smaller. The same pattern holds at the smaller scale: \textsc{ThinC}-1.7B reaches $42.8\%$, outperforming both Qwen3-1.7B ($32.2\%$), Qwen3-1.7B$^{*}$ ($29.8\%$), and CoRT-1.5B ($25.7\%$). Together, these results indicate that \textsc{ThinC} training yields consistent gains across scales beyond those obtained from tool-use prompting alone.

\paragraph{\textsc{ThinC} reasoning outperforms interleaved TIR.}
To isolate the effect of the reasoning format, we treat ASTER-4B 
as a natural ablation baseline for the interleaved approach. 
The two systems share a base model(Qwen3-4B-thinkning-2507), teacher capacity, and RL 
pipeline, differing primarily in trajectory structure. Under 
these matched conditions, \textsc{ThinC}-4B exceeds ASTER-4B 
on every benchmark by an average of $4.1$ points. The gain 
comes with lower inference cost, as \textsc{ThinC}-4B 
requires fewer tool calls per trajectory ($6.1$ vs.\ $11.1$) 
and produces shorter responses ($13.5$k vs.\ $15.4$k tokens; 
see Appendix~\ref{app:benchmark-analysis}). Code-centric reasoning therefore delivers higher accuracy over interleaved TIR, while naturally reducing inference overhead.

\subsection{SFT Cold-Start and RL Training Dynamics}
\label{sec:exp:dynamics}

\begin{figure}[t]
    \centering
    \includegraphics[width=\linewidth]{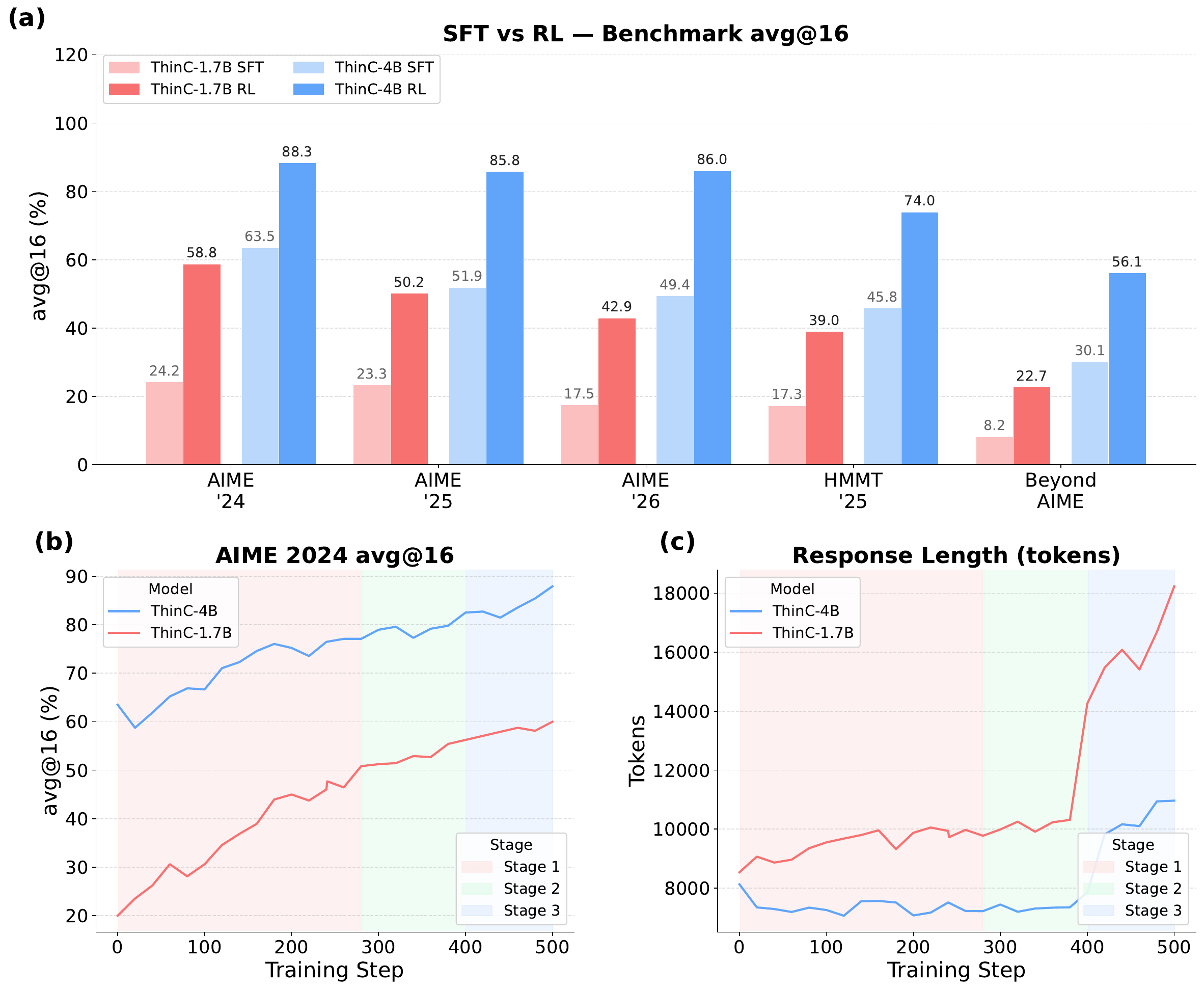}
    \caption{
  Training dynamics. (a) Benchmark avg@16 after SFT (light) 
  and after RL (dark) for \textsc{ThinC}-1.7B and 
  \textsc{ThinC}-4B across five math benchmarks. 
  (b) AIME 2024 avg@16 over RL training steps. 
  (c) Average response length over RL training steps. 
  Shaded regions in (b) and (c) denote the three RL stages.
}
    \label{fig:sft-rl-val}
\end{figure}

\paragraph{SFT establishes the format; RL drives the gains.}
After SFT, \textsc{ThinC}-4B-SFT reaches $48.1\%$ on average, below
both the teacher Qwen3.5-27B ($64.7\%$) and the tool-prompted
base model ($62.9\%$); \textsc{ThinC}-1.7B reaches $18.1\%$, also
below its base ($30.2\%$). This drop is by design: SFT teaches the model to reason in the \textsc{ThinC} format, not to maximize accuracy. RL produces the
benchmark gains, adding $29.9$ points at $4$B and $24.6$ points
at $1.7$B (Figure~\ref{fig:sft-rl-val}a) and lifting both
policies well above their bases and teachers.

\paragraph{RL improves the policy steadily throughout training.}
Figures~\ref{fig:sft-rl-val}b,c plot validation accuracy and
response length on AIME~2024 over RL steps. Both scales show
smooth, near-monotonic accuracy climbs with no plateau or
collapse, and the three-stage curriculum
(Section~\ref{sec:method:rl}) is visible as a mild inflection at
each stage boundary. Notably, \textsc{ThinC}-4B's response length
stays in the $7$K--$11$K range throughout RL, even when Stage~3
expands the context budget to $32$K. AIME~2024 accuracy
meanwhile climbs from $63.5\%$ at the SFT checkpoint to $88.3\%$ at
the end of RL. The $1.7$B model relies more on the extra
context, with response length roughly doubling in Stage~3.

\subsection{Does \textsc{ThinC} Actually Think in Code?}
\label{sec:exp:code-centric}

\begin{figure}[t]
    \centering
    \includegraphics[width=\linewidth]{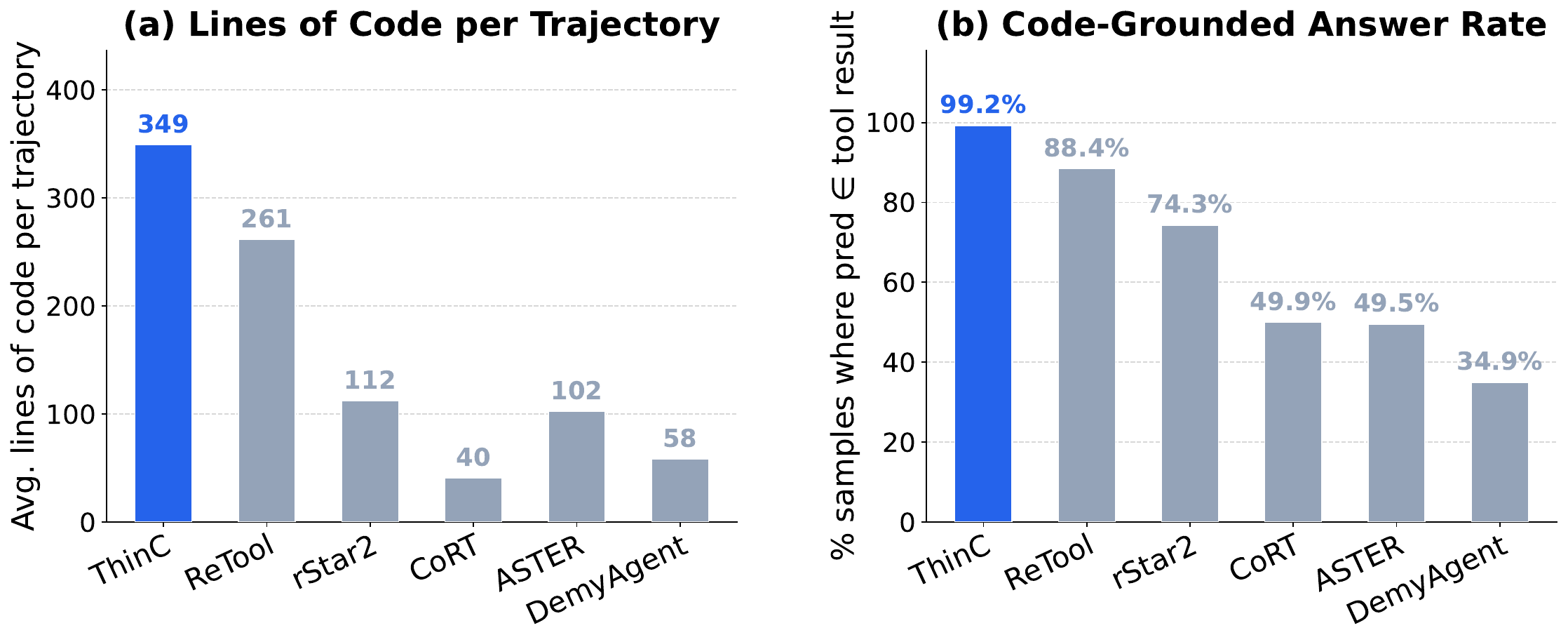}
    \caption{Code-centric reasoning behavior measured on overall benchmarks. (a) Average 
    lines of code per trajectory. (b) Code-grounded answer 
    rate, the fraction of trajectories whose final answer 
    appears in the output of some code block. 
    \textsc{ThinC}-4B leads on both metrics.}
    \label{fig:code-centric}
\end{figure}

We next verify that the trained model exhibits \textsc{ThinC} reasoning at inference time, as defined in Section~\ref{sec:method:reasoning}, rather than simply imitating the format of the training trajectories. Figure~\ref{fig:code-centric} compares \textsc{ThinC}-4B with five TIR baselines on AIME 2024--2026, HMMT 2025, and BeyondAIME using two complementary metrics. We consider both metrics together because they capture different aspects of code-centric reasoning. One measures how extensively code is used throughout the reasoning trajectory, while the other measures whether the final answer is grounded in interpreter outputs. Taken together, they provide a more complete view of whether a model not only writes code during reasoning, but also relies on execution outputs to produce its final answer.

\paragraph{\textsc{ThinC} shifts the reasoning process to code.}
\textsc{ThinC} writes an average of $349$ lines of code per sample (Figure~\ref{fig:code-centric}a), substantially more than ASTER ($102$), CoRT ($40$), and ReTool ($261$). Notably, ASTER and CoRT are the two TIR systems most explicitly designed to strengthen tool use, while ReTool is the strongest baseline on this metric. These results show that \textsc{ThinC} makes substantially heavier use of code throughout the reasoning trajectory. We next examine whether this code also serves as the primary driver of reasoning, rather than merely accompanying an NL derivation.

\paragraph{\textsc{ThinC} answers are grounded in code 
execution.}
We next examine whether the final answer of each trajectory 
appears in the execution output of at least one code block 
(Figure~\ref{fig:code-centric}b). \textsc{ThinC}-4B satisfies this condition in $99.2\%$ of trajectories, compared with $88.4\%$ for ReTool and $74.3\%$ for rStar2. Several other baselines are lower still, indicating that a large fraction of their final answers are generated through NL reasoning rather than code execution. As a result, they bypass the interpreter and remain vulnerable to the arithmetic and algebraic errors discussed in Section~\ref{sec:introduction}, where even a single NL mistake can corrupt the result. \textsc{ThinC} largely removes this failure mode by design: because its trajectory format contains no NL channel between code blocks, the final answer must be derived from interpreter output. Appendix~\ref{appendix:case-study} traces a representative 
\textsc{ThinC}-4B rollout that makes this pattern concrete. 
On AIME 2026 Problem 3, the model's $t_1$ contains only an 
algebraic restructuring of the problem 
($a+b+ab=(a{+}1)(b{+}1)-1$); the answer is then computed 
and cross-validated through multiple code-driven turns, with 
the model auditing and refining its own loop logic entirely 
within the next code block rather than via NL reasoning 
between blocks.

\subsection{Can \textsc{ThinC} Recover from Code Failures Without NL Reasoning?}

\textsc{ThinC}'s code-centric design raises a natural question 
about robustness: "With no NL reasoning between code blocks, what happens when a code execution fails?" Interleaved TIR can absorb the 
failure in NL reasoning and reframe the next attempt; \textsc{ThinC} 
cannot. Whether this hurts robustness or helps it is an empirical question we test here.

We measure this with \emph{Recovery@$k$}: among trajectories 
whose first $k$ code blocks all raise an interpreter error, the 
fraction that still arrives at the correct final answer. We 
compute the metric on AIME 2024--2026, HMMT 2025, and 
BeyondAIME, sweeping $k$ from $1$ to $5$.

\label{sec:analysis:recovery}
\begin{wrapfigure}{r}{0.55\linewidth}
    \centering
    \vspace{-0.8em}
    \includegraphics[width=\linewidth]{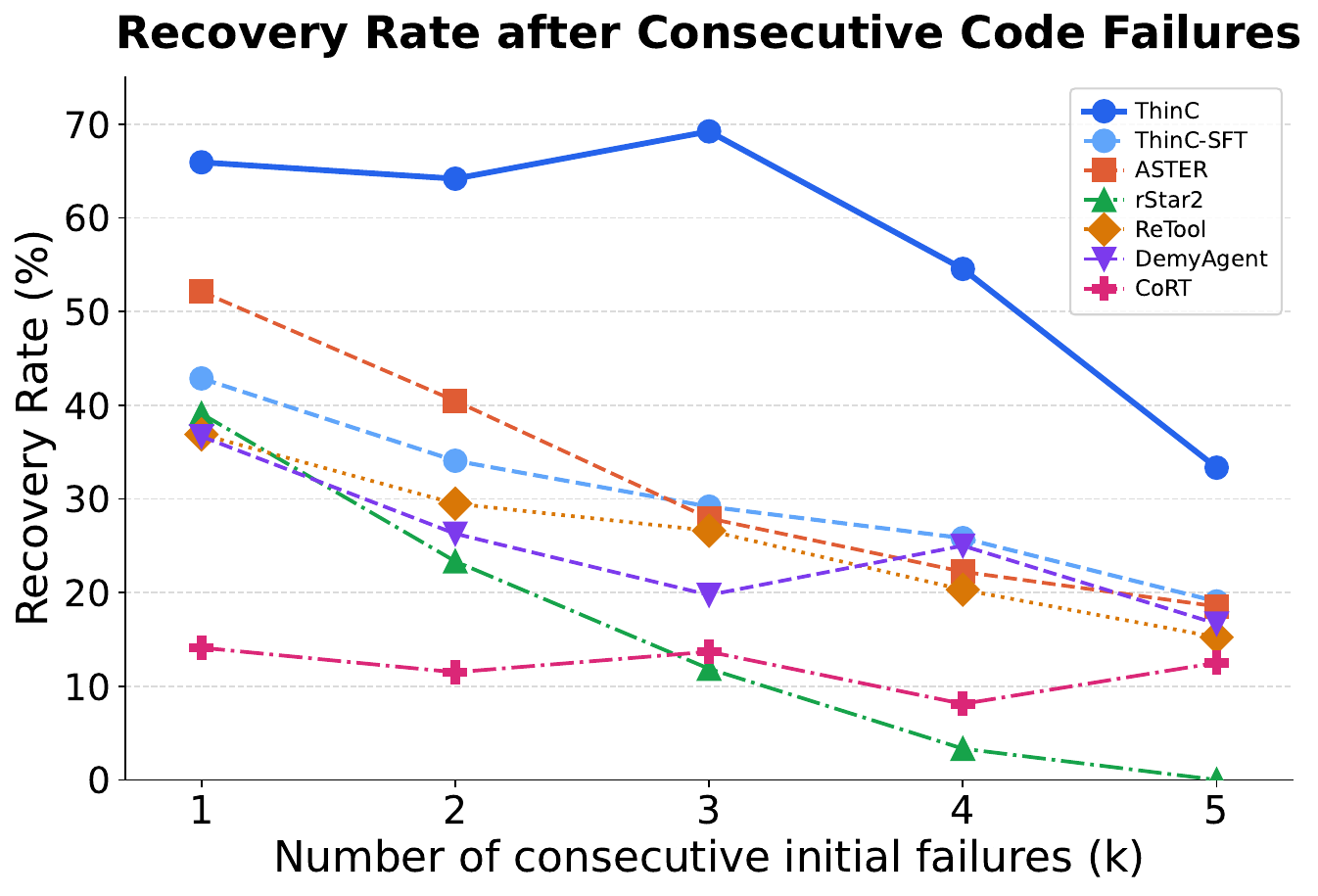}
    \caption{Recovery@$k$ under initial code failures. \textsc{ThinC} remains substantially more robust as early execution failures accumulate.}
    \label{fig:recovery}
    \vspace{-1.2em}
\end{wrapfigure}

\paragraph{Interleaved baselines degrade with $k$; \textsc{ThinC}-4B 
stays robust.}
Every interleaved TIR system loses ground as initial failures 
accumulate (Figure~\ref{fig:recovery}). ASTER drops from $52.1\%$ 
at $k=1$ to $18.5\%$ at $k=5$; rStar2-Agent collapses from $39.1\%$ 
to $0\%$; ReTool, DemyAgent, and CoRT decline along similar 
trajectories. \textsc{ThinC}, in contrast, stays in a narrow 
$64$--$69\%$ band across $k=1,2,3$, before declining to $54.5\%$ 
at $k=4$ and $33.3\%$ at $k=5$. Even at $k=5$, it recovers nearly 
$2\times$ as often as any interleaved baseline.

\paragraph{Partial robustness from the format, the rest from RL.}
Our SFT data is filtered to retain only trajectories that 
execute every code block without error, so recovery from 
failures is not directly demonstrated during SFT. Even so, 
\textsc{ThinC}-4B-SFT already exceeds most interleaved baselines, 
recovering in $42.9\%$ of trajectories at $k=1$ and $19.1\%$ at 
$k=5$. Only ASTER surpasses it, and only at $k\leq2$. The 
code-centric format, therefore, confers some robustness even 
before RL. RL on top of this prior lifts recovery by another 
$20$+ points across $k=1,2,3$, accounting for most of 
\textsc{ThinC}'s gap over the interleaved baselines.

\section{Discussion and Conclusion}
\label{sec:discussion}
We introduce \textsc{ThinC}, a code-centric approach to TIR in which the model reasons directly in code after a brief NL planning step. \textsc{ThinC}-4B reaches $78.1\%$ on five 
competition-level math benchmarks, outperforming all 
baselines. Two findings explain the gain: $99.2\%$ of final 
answers come from interpreter output rather than NL, and 
\textsc{ThinC} maintains $64$--$69\%$ recovery through three 
consecutive code failures where interleaved baselines degrade 
sharply. 

Several limitations remain. Due to computational constraints, 
our experiments are restricted to $1.7$B and $4$B parameters, 
and our evaluation is confined to competition-level 
mathematics. Whether \textsc{ThinC} scales to larger models, 
and whether the code-centric format extends to other 
tool-integrated reasoning domains, are natural directions for 
future work.


\ifincludebibliography
\bibliographystyle{plainnat}
\bibliography{references}
\fi

\newpage
\appendix
\UseRawInputEncoding
\lstdefinestyle{python}{
    language=Python,
    basicstyle=\ttfamily\footnotesize,
    keywordstyle=\color{blue},
    commentstyle=\color{gray}\itshape,
    stringstyle=\color{purple},
    backgroundcolor=\color{gray!5},
    frame=single,
    framerule=0pt,
    breaklines=true,
    showstringspaces=false,
    keepspaces=true,
}

\lstdefinestyle{prompt}{
    basicstyle=\ttfamily\footnotesize,
    backgroundcolor=\color{gray!5},
    frame=single,
    framerule=0pt,
    breaklines=true,
    breakatwhitespace=true,
    columns=fullflexible,
    showstringspaces=false,
    keepspaces=true,
    aboveskip=0.5em,
    belowskip=0.5em,
}

%
%
%

\definecolor{thinkbg}{HTML}{F4F7FB}
\definecolor{thinkrl}{HTML}{1F4E79}
\definecolor{codebg}{HTML}{F8F8F4}
\definecolor{codecmt}{HTML}{6A737D}
\definecolor{codekw}{HTML}{8959A8}
\definecolor{codestr}{HTML}{718C00}
\definecolor{resbg}{HTML}{F0F7F0}
\definecolor{resrl}{HTML}{2E6F2E}
\definecolor{ansbg}{HTML}{FFF6E5}
\definecolor{ansrl}{HTML}{B86E00}

\lstdefinestyle{thincpy}{%
  language=Python,
  basicstyle=\footnotesize\ttfamily,
  commentstyle=\color{codecmt}\itshape,
  keywordstyle=\color{codekw}\bfseries,
  stringstyle=\color{codestr},
  showstringspaces=false,
  breaklines=true,
  breakatwhitespace=true,
  columns=fullflexible,
  keepspaces=true,
  aboveskip=2pt, belowskip=2pt,
  xleftmargin=4pt, xrightmargin=4pt,
}

\newtcolorbox{thinkbox}{%
  enhanced, colback=thinkbg, colframe=thinkrl, coltitle=white,
  fonttitle=\sffamily\bfseries, title=\texttt{<think>},
  boxrule=0.4pt, arc=2pt, left=4pt, right=4pt, top=4pt, bottom=4pt,
  before skip=4pt, after skip=4pt,
}
\newtcolorbox{pyturn}[1]{%
  enhanced, colback=codebg, colframe=black!55, coltitle=white,
  fonttitle=\sffamily\bfseries, title=\texttt{<python>}~~Turn #1,
  boxrule=0.4pt, arc=2pt, left=2pt, right=2pt, top=2pt, bottom=2pt,
  before skip=4pt, after skip=2pt,
}
\newtcolorbox{resbox}{%
  enhanced, colback=resbg, colframe=resrl, coltitle=white,
  fonttitle=\sffamily\bfseries, title=\texttt{<result>},
  boxrule=0.4pt, arc=2pt, left=4pt, right=4pt, top=3pt, bottom=3pt,
  before skip=2pt, after skip=4pt, fontupper=\footnotesize\ttfamily,
}
\newtcolorbox{ansbox}{%
  enhanced, colback=ansbg, colframe=ansrl, coltitle=white,
  fonttitle=\sffamily\bfseries, title=\texttt{<answer>},
  boxrule=0.4pt, arc=2pt, left=4pt, right=4pt, top=3pt, bottom=3pt,
  before skip=4pt, after skip=4pt,
}

\section{Case Study: A \textsc{ThinC}-4B Trajectory on AIME 2026 Problem 3}
\label{appendix:case-study}

To illustrate the structural pattern of \textsc{ThinC}, we walk through one full rollout on
\textbf{AIME 2026 Problem 3}. The trajectory shows how a single brief
strategic plan in NL transitions into a sequence of code-centric
reasoning steps, each conditioned on the previous interpreter output,
with the final answer emerging from code execution rather than from
NL-level computation.

\subsection{Problem}

\begin{quote}\itshape
  Find the number of integers less than or equal to $100$ that are
  equal to $a+b+ab$ for some choice of distinct positive integers
  $a$ and $b$.
\end{quote}
The reference answer is $\boxed{70}$.

\subsection{Stage~1 --- Strategic Planning ($t_1$)}

The single \texttt{<think>} block contains no numerical computation.
It performs one algebraic restructuring of the problem---factoring
$a+b+ab=(a{+}1)(b{+}1)-1$---and specifies the resulting search problem
to be solved in code.

\begin{thinkbox}\small
I need to find the number of integers $\le 100$ that can be written as
$a+b+ab$ where $a$ and $b$ are distinct positive integers.

\medskip
Let me think about this expression: $a+b+ab=(a{+}1)(b{+}1)-1$.

So I need to find how many integers $n\le 100$ satisfy
$n=(a{+}1)(b{+}1)-1$ for some distinct positive integers $a,b$.
This means $n+1=(a{+}1)(b{+}1)$ where $a,b\ge 1$ and $a\ne b$.

\medskip
Let $m=a{+}1$, $n'=b{+}1$. Then $m,n'\ge 2$, $m\ne n'$, and
$m\cdot n'=n+1$. So I need to count how many values of $n\le 100$ can
be written as $m\cdot n'-1$ where $m,n'\ge 2$, $m\ne n'$, and
$m\cdot n'\le 101$.

\medskip
Actually, let me just brute force this. Since $a$ and $b$ are distinct
positive integers, let me iterate over possible values and check
which sums are $\le 100$. \textbf{Let me write code to compute this
directly.}
\end{thinkbox}

This realises the role \textsc{ThinC} reserves for $t_1$
(Section~\ref{sec:method:reasoning}): a high-level strategic plan,
not a step-by-step derivation. Three properties of this think block
illustrate the design intent:
\begin{enumerate}\itemsep2pt
  \item \textbf{No arithmetic in NL.} The model never evaluates an
    expression or enumerates a case manually. The think channel is
    used purely for symbolic restructuring of the problem.
  \item \textbf{Search-space specification.} It fixes the exact
    constraints (\,$m,n'\ge 2$, $m\ne n'$, $m\cdot n'\le 101$\,)
    before any code is written, providing a clean starting point for
    the code blocks that follow.
  \item \textbf{Single transition to code.} The block ends with one
    sentence (\textit{``Let me write code to compute this directly''})
    and from this point onward, NL does not return: every subsequent
    reasoning step is carried out inside a \texttt{<python>} block.
\end{enumerate}

\subsection{Stage~2 --- Code-Centric Reasoning}

The remainder of the rollout consists of five
\texttt{<python>}/\texttt{<result>} exchanges. Each code block builds
on the execution outputs of the preceding blocks, as prescribed by
Eq.~\ref{eq:thinc}, and every reasoning step---including
self-correction, structural verification, and independent
re-derivation---is performed in code.

\paragraph{Turn~1: Initial brute force.}
The model implements a direct enumeration with an early
\texttt{break} that it pre-emptively flags as potentially incorrect.

\begin{pyturn}{1}
\begin{lstlisting}[style=thincpy]
# (a+1)(b+1) - 1 -- brute force over distinct (a,b)
results = set()
for a in range(1, 100):
    for b in range(a+1, 100):     # b > a to keep pairs distinct
        n = a + b + a*b
        if n <= 100:
            results.add(n)
        elif n > 100:
            break                 # potentially premature -- revisited next turn
print(f"Number of distinct values: {len(results)}")
\end{lstlisting}
\end{pyturn}
\vspace{-3pt}
\begin{resbox}
Number of distinct values: 70
\end{resbox}

\paragraph{Turn~2: Self-correction in code.}
Conditioning on Turn~1's output, the model revisits its loop bound
and formalises the monotonicity argument ($n=a+b(1{+}a)$ is monotone
in $b$ for fixed $a$) directly as a code comment, then reruns with
the corrected condition. The audit and fix occur entirely within the
code lines; no NL reasoning between blocks is required.

\begin{pyturn}{2}
\begin{lstlisting}[style=thincpy]
# For fixed a, n = a + b(1+a) is monotone in b.
# So `break` is justified -- but only after we hit n > 100, not earlier.
results = set()
for a in range(1, 100):
    for b in range(1, 100):
        if a == b: continue
        n = a + b + a*b
        if n > 100: break
        results.add(n)
print(f"Number of distinct values: {len(results)}")
print("All values sorted:", sorted(results))
\end{lstlisting}
\end{pyturn}
\vspace{-3pt}
\begin{resbox}
Number of distinct values: 70
All values sorted: [5, 7, 9, 11, 13, 14, 15, 17, 19, 20, 21, 23, 25, 26, 27,
29, 31, 32, 33, 34, 35, 37, 38, 39, 41, 43, 44, 45, 47, 49, 50, 51, 53, 54,
55, 56, 57, 59, 61, 62, 63, 64, 65, 67, 68, 69, 71, 73, 74, 75, 76, 77, 79,
80, 81, 83, 84, 85, 86, 87, 89, 90, 91, 92, 93, 94, 95, 97, 98, 99]
\end{resbox}

\paragraph{Turn~3: Structural sanity check.} The model decomposes the
result by parity to confirm there is no off-by-one error in the
value set.

\begin{pyturn}{3}
\begin{lstlisting}[style=thincpy]
values = sorted(results)
odds  = [v for v in values if v % 2 == 1]
evens = [v for v in values if v % 2 == 0]
print(f"Count: {len(values)}")
print(f"Odd: {len(odds)}, Even: {len(evens)}")
\end{lstlisting}
\end{pyturn}
\vspace{-3pt}
\begin{resbox}
Count: 70
Odd: 48, Even: 22
\end{resbox}

\paragraph{Turn~4: Independent re-derivation.} The model re-implements
the count in the dual formulation $(u,v)=(a{+}1,b{+}1)$ and verifies
that it matches Turn~2.

\begin{pyturn}{4}
\begin{lstlisting}[style=thincpy]
# Method 2: enumerate (u, v) with u, v >= 2, u != v, u*v <= 101.
results2 = set()
for u in range(2, 102):
    for v in range(2, 102):
        if u == v: continue
        if u * v <= 101:
            n = u * v - 1
            if n <= 100:
                results2.add(n)
print(f"Method 2 count: {len(results2)}")
print(f"Same as method 1? {len(results2) == 70}")
\end{lstlisting}
\end{pyturn}
\vspace{-3pt}
\begin{resbox}
Method 2 count: 70
Same as method 1? True
\end{resbox}

\paragraph{Turn~5: Complement audit.} As a final empirical check, the
model lists the $30$ values in $\{1,\dots,100\}$ that are \emph{not}
attained.

\begin{pyturn}{5}
\begin{lstlisting}[style=thincpy]
missing = [i for i in range(1, 101) if i not in values]
print(f"Missing values (first 10): {missing[:10]}")
print(f"Total missing: {len(missing)}")
print(f"Total present: {len(values)}")
\end{lstlisting}
\end{pyturn}
\vspace{-3pt}
\begin{resbox}
Missing values (first 10): [1, 2, 3, 4, 6, 8, 10, 12, 16, 18]
Total missing: 30
Total present: 70
\end{resbox}

\begin{ansbox}\small
\centering
\texttt{<answer>} The final answer is $\boxed{70}$ \texttt{</answer>}
\end{ansbox}

\subsection{What This Trajectory Illustrates}

The rollout exhibits, in a single sample, the four behaviours that
the \textsc{ThinC} format establishes by construction:

\begin{description}\itemsep3pt
  \item[\textbf{Strategy-only $t_1$.}] The think channel carries a
    single algebraic insight ($a+b+ab=(a{+}1)(b{+}1)-1$) and the
    search constraints, with no arithmetic performed in NL. This
    realises the \textsc{ThinC} constraint that $t_1$ expresses
    strategy rather than derivation
    (Section~\ref{sec:method:reasoning}).
  \item[\textbf{Code as the reasoner.}] Every reasoning step from
    Turn~1 onward is performed in code. Self-correction (Turn~2),
    structural verification (Turn~3), independent re-derivation
    (Turn~4), and complement audit (Turn~5) all occur within
    \texttt{<python>} blocks; NL appears only as inline code comments
    and never between code blocks.
  \item[\textbf{Conditioning on execution outputs.}] Turn~2's repair
    is grounded in the model's audit of Turn~1's output. Each
    subsequent turn similarly conditions on $o_{<i}$, consistent with
    the formal trajectory structure in Eq.~\ref{eq:thinc}.
  \item[\textbf{Code-grounded final answer.}] The committed answer
    ($70$) appears in the interpreter output of Turns~2, $4$, and
    $5$, rather than being generated in NL.
\end{description}

This rollout is the qualitative correlate of the quantitative gap
between \textsc{ThinC}-4B and interleaved TIR baselines in
Table~\ref{tab:main} and the $99.2\%$ code-grounded answer rate
reported in Section~\ref{sec:exp:code-centric}: the policy has learned
to confine NL to strategic planning and to carry out all subsequent
reasoning in code.

\section{Few-Shot Prompt for Trajectory Distillation}
\label{app:prompt}
\subsection{Few-Shot Prompt for Trajectory Distillation}

This appendix presents the full $3$-shot prompt used to elicit 
\textsc{ThinC} trajectories from the teacher model 
(Qwen3.5-27B). 

\begin{tcolorbox}[
  breakable, 
  enhanced, 
  colback=gray!5, 
  colframe=gray!40, 
  boxrule=0.5pt,
  title={Few-shot prompt}
]
\begin{lstlisting}[style=prompt]
You are a math problem solver that works exclusively through Python 
code execution.

Rules:
1. Write Python code inside <python> </python> tags. You will receive 
   execution results inside <result> </result> tags.
2. Each code block runs in a separate process. Variables do NOT 
   persist between turns. Re-define or hardcode values from previous 
   results.
3. You may think freely before the first code block, but your final output 
   to the user must be ONLY <python> blocks and one <answer> block.
4. Embed your reasoning as concise comments inside the code. Each 
   code block should contain short comments explaining: what you are 
   computing and why, based on previous results.
5. Solve problems step by step: compute intermediate results in each 
   code block, observe the output, then decide what to compute next 
   based on those results. Each block should advance the solution by 
   one logical step.
6. When you have the final answer, provide it inside 
   <answer> The final answer is \boxed{answer} </answer>.

CRITICAL: Each code block is executed in a completely fresh Python 
process. You MUST re-import libraries and re-define all variables 
in every code block. Use hardcoded values from previous <result> 
outputs, not variable names.

Here are some examples:

Question: Find the smallest positive integer $k$ with the following property: if each cell of a $100\\times 100$ grid is dyed with one color and the number of cells of each color is not more than $104$, then there is a $k\\times1$ or $1\\times k$ rectangle that contains cells of at least three different colors.
Output:
<python>
# 100x100 grid, each color used at most 104 times
# Minimum number of distinct colors required by pigeonhole
import math
total_cells = 100 * 100
max_per_color = 104
min_colors = math.ceil(total_cells / max_per_color)
print(f"total cells: {total_cells}, min colors needed: {min_colors}")
</python>
<result>
total cells: 10000, min colors needed: 97
</result>
<python>
# If every k consecutive cells in a row have <= 2 colors,
# colors must form contiguous blocks.
# Intermediate blocks need width >= k-1 to prevent a k-window spanning 3 blocks.
# So max colors per row = floor((100 - 2) / (k-1)) + 2
# (first and last block can be width 1, each intermediate block >= k-1)
for k in range(10, 15):
    max_colors_per_row = (100 - 2) // (k - 1) + 2
    print(f"k={k}: max colors per row = {max_colors_per_row}")
</python>
<result>
k=10: max colors per row = 12
k=11: max colors per row = 11
k=12: max colors per row = 10
k=13: max colors per row = 10
k=14: max colors per row = 9
</result>
<python>
# Construction attempt: divide grid into s x s blocks, one color per block.
# Block size s >= k-1 satisfies both row and column constraints.
# Each color occupies s^2 cells, need s^2 <= 104.
import math
for s in range(9, 13):
    num_colors = math.ceil(100 / s) ** 2
    cells_per_color = s * s
    print(f"block {s}x{s}: colors={num_colors}, cells_per_color={cells_per_color}, feasible={cells_per_color <= 104}")
</python>
<result>
block 9x9: colors=144, cells_per_color=81, feasible=True
block 10x10: colors=100, cells_per_color=100, feasible=True
block 11x11: colors=100, cells_per_color=121, feasible=False
block 12x12: colors=81, cells_per_color=144, feasible=False
</result>
<python>
# s=10 works (100 <= 104). Verify: build 10x10 block grid,
# check every 11-window in all rows and columns has <= 2 colors
from collections import Counter
grid = [[0]*100 for _ in range(100)]
color = 0
for bi in range(10):
    for bj in range(10):
        for r in range(bi*10, (bi+1)*10):
            for c in range(bj*10, (bj+1)*10):
                grid[r][c] = color
        color += 1
counts = Counter(grid[r][c] for r in range(100) for c in range(100))
# check all 11-windows in rows
row_ok = all(len(set(grid[r][c:c+11])) <= 2 for r in range(100) for c in range(90))
# check all 11-windows in columns
col_ok = all(len(set(grid[r2][c] for r2 in range(r, r+11))) <= 2 for c in range(100) for r in range(90))
print(f"k=11 row: {'PASS' if row_ok else 'FAIL'}, col: {'PASS' if col_ok else 'FAIL'}")
print(f"colors: {len(counts)}, max per color: {max(counts.values())}")
</python>
<result>
k=11 row: PASS, col: PASS
colors: 100, max per color: 100
</result>
<python>
# Same 10x10 blocks: check k=12. Block size 10 < 12-1=11,
# so a 12-window crossing two block boundaries will see 3 colors.
grid = [[0]*100 for _ in range(100)]
color = 0
for bi in range(10):
    for bj in range(10):
        for r in range(bi*10, (bi+1)*10):
            for c in range(bj*10, (bj+1)*10):
                grid[r][c] = color
        color += 1
found = None
for r in range(100):
    for c in range(89):
        if len(set(grid[r][c:c+12])) >= 3:
            found = (r, c, grid[r][c:c+12])
            break
    if found:
        break
print(f"k=12 window: row={found[0]}, start={found[1]}, cells={found[2]}")
</python>
<result>
k=12 window: row=0, start=9, cells=[0, 1, 1, 1, 1, 1, 1, 1, 1, 1, 1, 2]
</result>
<python>
# For k=12, any block coloring needs both dimensions >= 11.
# Exhaustively check: no rectangle with both dims >= 11 has area <= 104.
import math
feasible = []
for p in range(11, 105):
    for q in range(11, 105):
        if p * q <= 104:
            feasible.append((p, q, math.ceil(100/p) * math.ceil(100/q)))
if feasible:
    for p, q, nc in feasible:
        print(f"{p}x{q}: colors={nc}")
else:
    print(f"no block with both dims >= 11 fits in 104 (min 11x11={11*11})")
</python>
<result>
no block with both dims >= 11 fits in 104 (min 11x11=121)
</result>
<answer> The final answer is \\boxed{12} </answer>

Question: In the Cartesian coordinate system \\( xOy \\), the area of the region corresponding to the set of points \\( K = \\{(x, y) \\mid (|x| + |3y| - 6)(|3x| + |y| - 6) \\leq 0 \\} \\) is ________.
Output:
<python>
# (f1)(f2) <= 0 means exactly one of f1<=0, f2<=0
# Region A: |x|+3|y|<=6 -> |x|/6+|y|/2<=1, area = 2*a*b = 2*6*2
# Region B: |3x|+|y|<=6 -> |x|/2+|y|/6<=1, area = 2*a*b = 2*2*6
from sympy import Rational, solve, symbols
x, y = symbols('x y')
area_A = 2 * 6 * 2
area_B = 2 * 2 * 6
print(f"area A: {area_A}, area B: {area_B}")
</python>
<result>
area A: 24, area B: 24
</result>
<python>
# Find where the two boundaries intersect in the first quadrant
# x+3y=6 and 3x+y=6
from sympy import symbols, solve
x, y = symbols('x y')
sol = solve([x + 3*y - 6, 3*x + y - 6], [x, y])
print(f"intersection of boundaries: {sol}")
</python>
<result>
intersection of boundaries: {x: 3/2, y: 3/2}
</result>
<python>
# First quadrant intersection region A∩B has vertices:
# (0,0), (2,0) from |3x|+|y|=6, (3/2,3/2) from boundary crossing, (0,2) from |x|+3|y|=6
# Compute area via shoelace, then multiply by 4 for all quadrants
from sympy import Rational
pts = [(0,0), (2,0), (Rational(3,2), Rational(3,2)), (0,2)]
n = len(pts)
s = 0
for i in range(n):
    j = (i+1) % n
    s += pts[i][0]*pts[j][1] - pts[j][0]*pts[i][1]
area_q1 = abs(s) / 2
area_intersection = 4 * area_q1
print(f"first quadrant: {area_q1}, total intersection: {area_intersection}")
</python>
<result>
first quadrant: 3, total intersection: 12
</result>
<python>
# K = symmetric difference of A and B = A + B - 2*(A∩B)
area_A = 24
area_B = 24
area_intersection = 12
area_K = area_A + area_B - 2 * area_intersection
print(f"area K = {area_K}")
</python>
<result>
area K = 24
</result>
<answer> The final answer is \\boxed{24} </answer>

Question: The area of an equilateral triangle ABC is 36. Points P, Q, R are located on BC, AB, and CA respectively, such that BP = 1/3 BC, AQ = QB, and PR is perpendicular to AC. Find the area of triangle PQR.
Output:
<python>
# Equilateral triangle area = (sqrt(3)/4)*s^2 = 36, solve for side length
import sympy as sp
s = sp.Symbol('s', positive=True)
side = sp.solve(sp.sqrt(3)/4 * s**2 - 36, s)
print(f"side: {side[0]}, side^2: {sp.simplify(side[0]**2)}")
</python>
<result>
side: 4*3**(3/4), side^2: 48*sqrt(3)
</result>
<python>
# Place triangle: B at origin, C at (a,0), A at (a/2, a*sqrt(3)/2)
# P on BC with BP = 1/3 BC, Q is midpoint of AB (AQ = QB)
import sympy as sp
from sympy import sqrt, Rational, Matrix
a = sp.Symbol('a', positive=True)
B = Matrix([0, 0])
C = Matrix([a, 0])
A = Matrix([a/2, a*sqrt(3)/2])
P = B + Rational(1,3) * (C - B)
Q = (A + B) / 2
print(f"P = {P.T}")
print(f"Q = {Q.T}")
</python>
<result>
P = Matrix([[a/3, 0]])
Q = Matrix([[a/4, sqrt(3)*a/4]])
</result>
<python>
# R on CA: R = A + t*(C - A). Find t such that PR is perpendicular to AC.
import sympy as sp
from sympy import sqrt, Rational, Matrix
a, t = sp.symbols('a t', positive=True)
A = Matrix([a/2, a*sqrt(3)/2])
C = Matrix([a, 0])
P = Matrix([a/3, 0])
R = A + t * (C - A)
AC = C - A
PR = R - P
# PR · AC = 0 for perpendicularity
t_val = sp.solve(PR.dot(AC), t)[0]
R_val = A + t_val * (C - A)
print(f"t = {t_val}")
print(f"R = {sp.simplify(R_val).T}")
</python>
<result>
t = 2/3
R = Matrix([[5*a/6, sqrt(3)*a/6]])
</result>
<python>
# Area of PQR using shoelace formula with P=(a/3,0), Q=(a/4,sqrt(3)a/4), R=(5a/6,sqrt(3)a/6)
import sympy as sp
from sympy import sqrt, Abs
a = sp.Symbol('a', positive=True)
P = (sp.Rational(1,3)*a, 0)
Q = (sp.Rational(1,4)*a, sqrt(3)/4*a)
R = (sp.Rational(5,6)*a, sqrt(3)/6*a)
area = Abs(P[0]*(Q[1]-R[1]) + Q[0]*(R[1]-P[1]) + R[0]*(P[1]-Q[1])) / 2
area = sp.simplify(area)
print(f"area PQR = {area}")
</python>
<result>
area PQR = 5*sqrt(3)*a**2/72
</result>
<python>
# Substitute a^2 = 48*sqrt(3) from the first step
import sympy as sp
area_PQR = 5 * sp.sqrt(3) * 48 * sp.sqrt(3) / 72
print(f"area PQR = {sp.simplify(area_PQR)}")
</python>
<result>
area PQR = 10
</result>
<answer> The final answer is \\boxed{10} </answer>

\end{lstlisting}
\end{tcolorbox}

\section{Tool Call and Response Length per Benchmark}
\label{app:benchmark-analysis}

This section reports the average number of tool calls per 
trajectory and the average response length on each 
evaluation benchmark.

\subsection{Tool Calls by Benchmark}
\label{app:toolcalls}

\begin{figure}[h]
    \centering
    \includegraphics[width=0.95\linewidth]{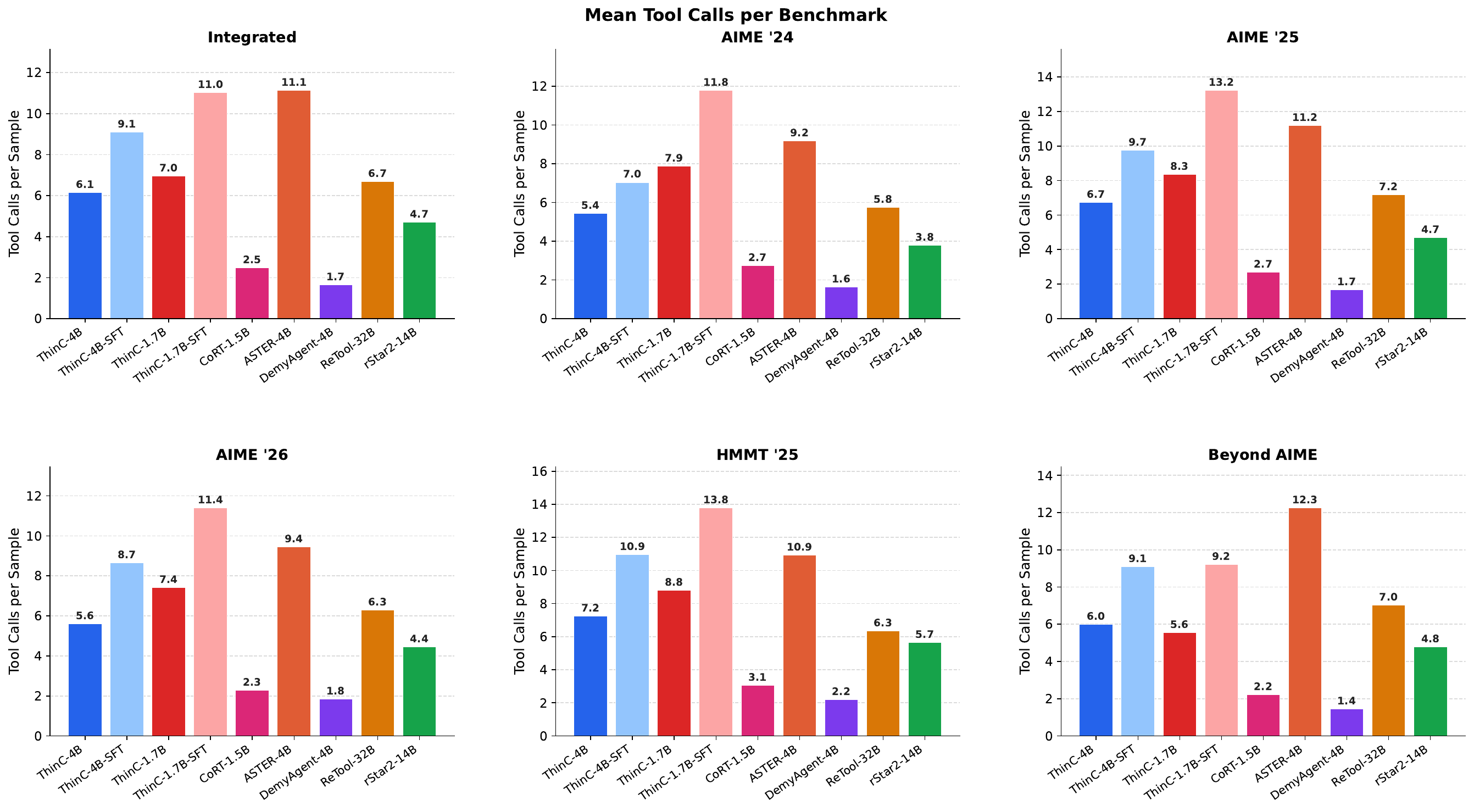}
    \caption{Average tool calls per benchmark. We compare how often models invoke the Python interpreter across AIME 2024--2026, HMMT 2025 February, and BeyondAIME.}
    \label{fig:toolcalls-benchmark}
\end{figure}

\subsection{Response Length by Benchmark}
\label{app:length}

\begin{figure}[h]
    \centering
    \includegraphics[width=0.95\linewidth]{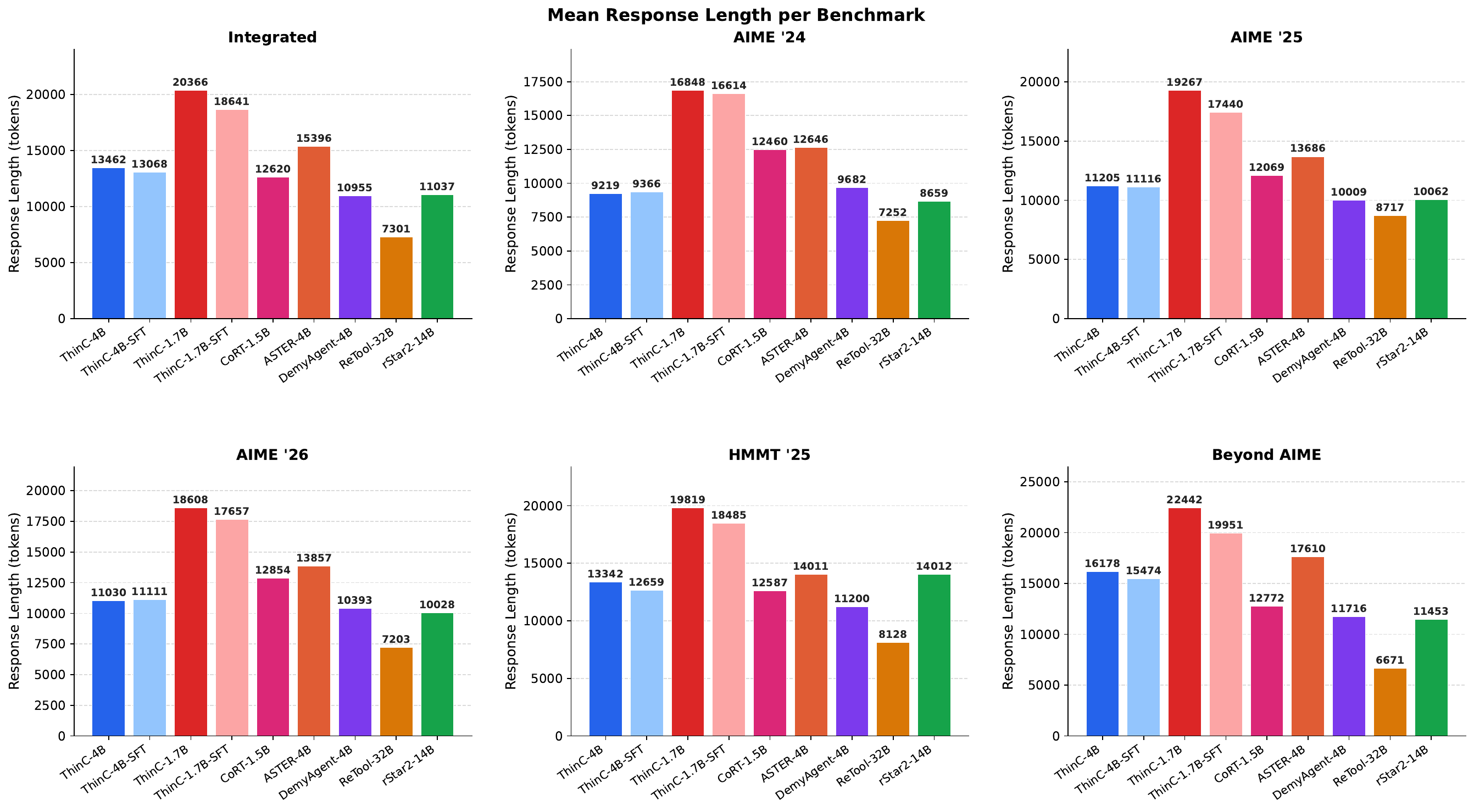}
    \caption{Average response length per benchmark. We report the mean trajectory length across AIME 2024--2026, HMMT 2025 February, and BeyondAIME. }
    \label{fig:length-benchmark}
\end{figure}

\section{OOD Generalization}
\label{app:ood}

We evaluate \textsc{ThinC}-4B on GPQA-Diamond~\citep{rein2024gpqa}, 
a graduate-level science benchmark covering physics, chemistry, 
and biology, as an out-of-distribution (OOD) test. 
Table~\ref{tab:gpqa} reports avg@16 and best@16 accuracy. 
\textsc{ThinC}-4B leads on both metrics, exceeding ASTER-4B by 
$3.1$ points on avg@16 and the base model by $7.6$ points on 
best@16, suggesting that the code-centric reasoning format 
generalizes more effectively than the interleaved alternative 
beyond the mathematical training domain.

\begin{table}[h]
    \centering
    \caption{GPQA-Diamond accuracy (\%). Best per column in 
    \textbf{bold}.}
    \label{tab:gpqa}
    \begin{tabular}{lcc}
    \toprule
    Model & \textbf{avg@16} & \textbf{best@16} \\
    \midrule
    Qwen3-4B-Thinking       & 66.32          & 83.84          \\
    ASTER-4B                & 63.42          & 90.40          \\
    \textsc{ThinC}-4B       & \textbf{66.48} & \textbf{91.41} \\
    \bottomrule
    \end{tabular}
\end{table}

\section{Training Details}
\label{app:training}

All training runs use a single node with $8\times$ NVIDIA H200 GPUs. 
We perform SFT using the LLaMA-Factory framework~\citep{zheng2024llamafactory} 
and RL using the verl framework~\citep{sheng2024hybridflow}. Both 
\textsc{ThinC}-1.7B and \textsc{ThinC}-4B share identical 
hyperparameter configurations at each training stage.

\subsection{Supervised Fine-Tuning}
\label{app:training:sft}

\begin{table}[h]
\centering
\caption{SFT hyperparameters for \textsc{ThinC}-1.7B-SFT and 
\textsc{ThinC}-4B-SFT.}
\label{tab:sft_hparams}
\begin{tabular}{ll}
\toprule
\textbf{Hyperparameter} & \textbf{Value} \\
\midrule
Per-device batch size       & $1$ \\
Gradient accumulation steps & $2$ \\
Global batch size           & $16$ \\
Learning rate               & $7 \times 10^{-6}$ \\
LR scheduler                & Cosine \\
Warmup ratio                & $0.1$ \\
Number of epochs            & $3$ \\
Context length              & $32{,}768$ \\
Precision                   & bfloat16 \\
\bottomrule
\end{tabular}
\end{table}

\subsection{Reinforcement Learning}
\label{app:training:rl}

Following DAPO~\citep{yu2025dapoopensourcellmreinforcement}, we 
apply token-level loss normalization across the rollout group, 
asymmetric clipping with $\epsilon_{\mathrm{low}} = 0.20$ and 
$\epsilon_{\mathrm{high}} = 0.28$, and disable the KL penalty. 
Table~\ref{tab:rl_hparams} lists the full hyperparameter 
configuration along with the three-stage curriculum.

\begin{table}[h]
\centering
\caption{RL hyperparameters for \textsc{ThinC}-1.7B and 
\textsc{ThinC}-4B.}
\label{tab:rl_hparams}
\begin{tabular}{ll}
\toprule
\textbf{Hyperparameter} & \textbf{Value} \\
\midrule
Algorithm                       & GRPO \\
Learning rate                   & $1 \times 10^{-6}$ \\
Loss normalization              & Token-level \\
$\epsilon_{\mathrm{low}}$       & $0.20$ \\
$\epsilon_{\mathrm{high}}$      & $0.28$ \\
KL coefficient                  & $0$ \\
Rollout prompt batch size       & $128$ \\
Trajectories per prompt ($G$)   & $8$ \\
\midrule
\multicolumn{2}{l}{\textit{Stage 1}} \\
\quad Context length            & $16{,}384$ \\
\quad Max tool calls            & $20$ \\
\quad Problem set               & Full DAPO-Math-17k \\
\midrule
\multicolumn{2}{l}{\textit{Stage 2}} \\
\quad Context length            & $16{,}384$ \\
\quad Max tool calls            & $20$ \\
\quad Problem set               & Stage 1 set with $100\%$-pass-rate problems removed \\
\midrule
\multicolumn{2}{l}{\textit{Stage 3}} \\
\quad Context length            & $32{,}768$ \\
\quad Max tool calls            & $40$ \\
\quad Problem set               & Same as Stage 2 \\
\bottomrule
\end{tabular}
\end{table}

\end{document}